\documentclass[preprint,12pt]{elsarticle}

\usepackage{amssymb}
\usepackage{amssymb,amsmath}
\usepackage{wrapfig}
\usepackage{graphicx,grffile}
\usepackage{lscape}
\usepackage{natbib}
\usepackage{adjustbox}
\usepackage{multirow}
\usepackage[T1]{fontenc} 
\usepackage{booktabs}
\usepackage{xurl}
\usepackage{xcolor}
\usepackage{rotating}

\journal{Journal of Biomedical Informatics}

\begin{document}

\begin{frontmatter}



\title{Progress Note Understanding - Assessment and Plan Reasoning: Overview of the 2022 N2C2 Track 3 Shared Task}


\author[inst1]{Yanjun Gao, PhD \corref{cor1}%
\fnref{fn1}}
\author[inst2]{Dmitriy Dligach, PhD}
\author[inst3]{Timothy Miller, PhD}
\author[inst1]{Matthew M. Churpek MD, MPH, PhD}
\author[inst4]{Ozlem Uzuner, PhD} 
\author[inst1]{Majid Afshar, MD, MSCR }

\address[inst1]{ICU Data Science Lab, Department of Medicine, University of Wisconsin Madison}
\address[inst2]{Department of Computer Science, Loyola University Chicago}
\address[inst3]{Boston Children’s Hospital, Harvard University}
\address[inst4]{Department of Information Sciences and Technology, George Mason University}

\cortext[cor1]{Corresponding author}
\fntext[fn1]{Email author: ygao@medicine.wisc.edu}

            
            
            

\begin{abstract}
Daily progress notes are a common note type in the electronic health record (EHR) where healthcare providers document the patient's daily progress and treatment plans. The EHR is designed to document all the care provided to patients, but it also enables note bloat with extraneous information that distracts from the diagnoses and treatment plans. Applications of natural language processing (NLP) in the EHR is a growing field with the majority of methods in information extraction. Few tasks use NLP methods for downstream diagnostic decision support. We introduced the 2022 National NLP Clinical Challenge (N2C2) Track 3: Progress Note Understanding - Assessment and Plan Reasoning as one steps towards a new suite of tasks. The Assessment and Plan Reasoning task focuses on the most critical components of progress notes, Assessment and Plan subsections where health problems and diagnoses are contained. The goal of the task was to develop and evaluate NLP systems that automatically predict causal relations between the overall status of the patient contained in the Assessment section and its relation to each component of the Plan section which contains the diagnoses and treatment plans. The goal of the task was to identify and prioritize diagnoses as the first steps in diagnostic decision support to find the most relevant information in long documents like daily progress notes. We present the results of the 2022 n2c2 Track 3 and provide a description of the data, evaluation, participation and system performance.          
\end{abstract}



\begin{keyword}
Natural language processing \sep clinical reasoning \sep clinical diagnostic decision support \sep national nlp clinical challenge
\end{keyword}

\end{frontmatter}



\section{Introduction}\label{sec1}
Healthcare providers generate notes in the electronic health record (EHR) to update diagnoses and treatment plans, and to document changes in the patient’s health status using the daily progress note type. The progress note is one of the most frequent note types that carry the most relevant and viewed documentation of the patient's care~\cite{brown2014physicians}. The Subjective, Objective, Assessment and Plan (SOAP) format is the framework currently taught in medical schools for generating a daily progress note and it remains the most ubiquitous format for daily care note taking in the EHR~\cite{weed1964medical}. The Assessment and Plan sections of the progress notes are the free-text fields where healthcare providers identify patients’ problems/diseases and treatment plans. Specifically, the Assessment section summarizes the patients’ active health problems or diseases for that day. The Plan section consists of multiple subsections, each addressing a specific health issue or diagnosis followed by a detailed treatment plan (e.g., Respiratory failure with MRSA pneumonia: continue seven days of vancomycin, continue mechanical ventilation with lung-protective strategy, wean oxygen). Patients typically have a main diagnosis or problem with associated conditions or downstream health effects. Healthcare providers can implicitly understand what parts of a treatment plan directly target the primary concerns of a series of health problems based on their background knowledge and medical reasoning. Still, this information is not easily available for downstream analysis. Automatic extraction of this information can potentially be useful for downstream use cases such as problem list generation.

However, EHRs suffer from issues such as note bloat (copying and pasting), information overload (automatic inclusion of data and administrative documentation), and disorganized notes, which can lead to burnout among providers and hinder efficient care~\cite{shoolin2013association}. Methods in Natural Language Processing (NLP) hold promise to help overcome the negative consequences of large-scale EHRs. More shared tasks are needed to focus on clinical decision support to guide providers through the breadth and depth of EHR data to reduce diagnostic errors and provide more efficient documentation. To date, tasks proposed for clinical NLP (cNLP) included named entity recognition (NER), information extraction (IE), relation extraction, document classification, sentence classification and others. In a scoping review on publicly available clinical NLP tasks from English sources between 2006 and 2021 showed that half of the tasks were for IE and NER, and over sixty percent of the tasks were non-specific secondary use in clinical settings~\cite{gao2022scoping}. While tasks and models for clinical IE are important, few tasks were motivated by providers' needs in daily practice and framed to assist the providers' decision-making~\cite{lederman2022tasks}. A shift towards more tasks designed for clinical decision support  (CDS) are needed~\cite{lederman2022tasks, gao2022scoping}.

To facilitate this paradigm shift and research focus of cNLP tasks for clinical decision-making, in particular, diagnostic decision-making, we prepared Track 3 of the 2022 National NLP Clinical Challenges (N2C2), a shared task named as \textsc{Progress Note Understanding - Assessment and Plan Reasoning}. Since 2006, the N2C2 has been a key contributor to the advancement of cNLP through shared tasks that focused on developing and evaluating NLP algorithms for extracting and organizing clinical data, including the EHR. We formulated the providers’ reasoning process as a relation prediction task between Assessment and every Plan Subsection within a single daily progress note. Given each pair of Assessment and Plan subsection as input, the goal was to develop a model that accurately predicted one of the following relations: \textsc{(1) Direct}, \textsc{ (2) Indirect}, \textsc{ (3) Neither} and \textsc{ (4) Not Relevant}. The four relations corresponded to the provders' judgment on whether a diagnosis presented in the Plan Subsection was the main reason for hospitalization (\textsc{Direct}), the secondary health problem/diagnosis to the main problem/diagnosis (\textsc{Indirect}), an issue that was not documented (\textsc{Neither}), and not a diagnosis or problem (\textsc{Not Relevant}). A model developed on this task could assist in prioritizing health concerns and allow guidance on the most critical and potentially life-threatening issues in a patient's medical chart. 

Predicting the relation between the Assessment and Plan sections helps to evaluate if the NLP model could use medical knowledge to perform reasoning. The aim of this paper was to present an overview of the shared task including the data preparation, evaluation, and submitted system performance. 

\section{Methods}
\subsection{Data Preparation}

Our task was based on the daily progress notes sampled from the Medical Information Mart for Intensive Care (MIMIC-III) dataset. MIMIC-III is a large, single-center EHR database with de-identified patient records in critical care units (ICU). It is publicly available through PhysioNet with a Data Usage Agreement. We randomly sampled a subset of 5000 physician-written progress notes over 84 note types that represent daily progress notes. The sampling method was multi-disciplinary and included notes from the Trauma ICU, Medical ICU, Surgical ICU, Cardiovascular ICU, Cardiothoracic ICU, and Neurological ICU. An initial screening was performed to exclude the progress notes that did not contain any diagnoses in the Assessment and Plan sections.

The task of \textsc{Assessment and Plan Reasoning} was the second task in a new suite of clinical NLP tasks, \textsc{Progress Note Understanding}~\cite{gao-etal-2022-hierarchical}. The suite aims at training and evaluating future NLP models for clinical text understanding, clinical knowledge representation, inference and summarization. To build this suite, we used an annotation guideline developed by two physicians with board certifications in critical care medicine and clinical informatics. The guidelines specified three sequential stages of the annotation process: \textsc{SOAP Section Tagging} labeling all sections of the progress note into a SOAP category; \textsc{Assessment and Plan Relation Labeling} specifying the relations between symptoms and problems
mentioned in the Assessment and diagnoses covered in
each Plan Subsection; \textsc{Problem List Identification}
highlighting the final diagnoses and treatment plans. The data for our N2C2 task comes from the second stage of the annotation \textsc{Assessment and Plan Relation Labeling}. The annotation guideline that described the rules for labelling the Plan subsections is presented in Figure~\ref{fig:guidelines}. 

\begin{figure}[ht]
    \centering
    \includegraphics[width=\textwidth]{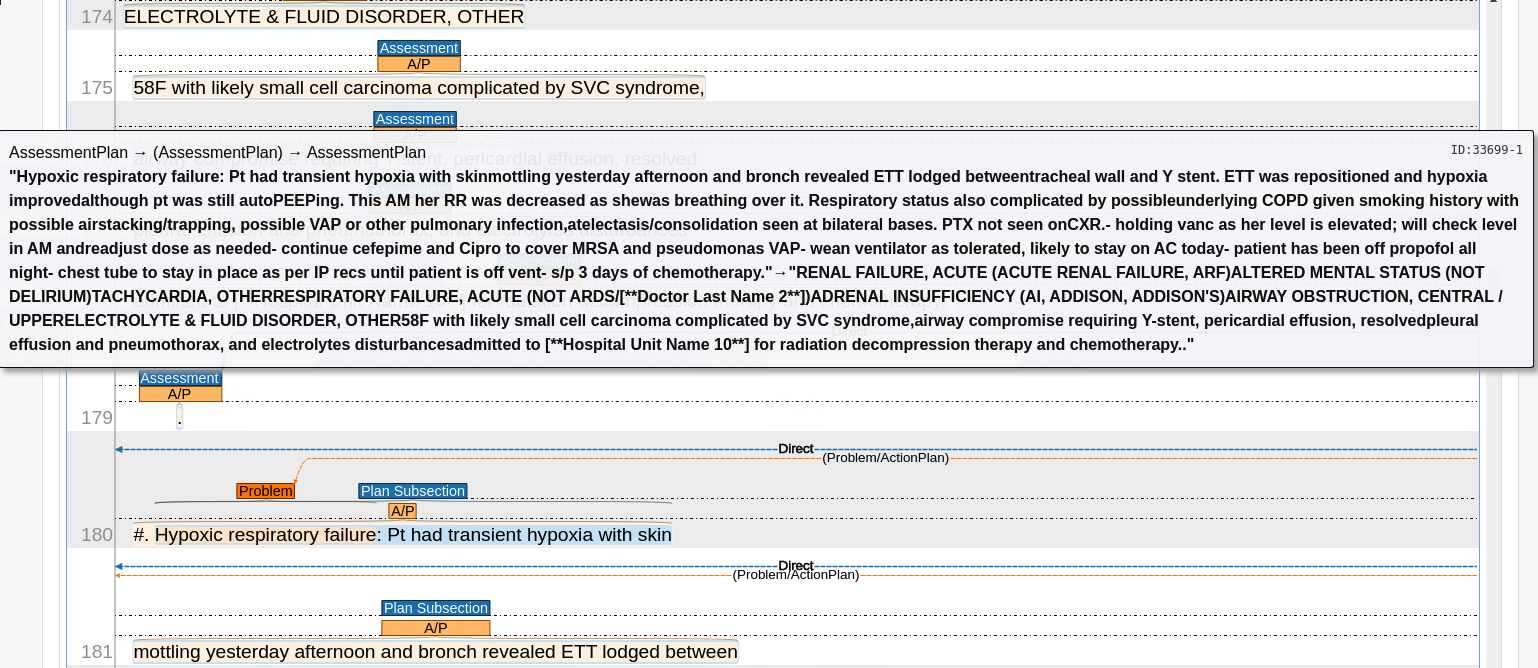}
    \caption{An example of INCEpTION interface for \textsc{Assessment and Plan Relation Labeling}. Linebreaks are naturally preserved from MIMIC-III raw notes. The pop-up window shows an link between the Plan Subsection ``Hypoxic respiratory failure: Pt ... 3 days of chemontherapy'' and the Assessment ``RENAL FAILURE, ACUTE 
 ...''. The relation label is \textsc{Direct}. }
    \label{fig:inception}
\end{figure}

We used INCEpTION~\cite{tubiblio106270} as the annotation platform. Figure~\ref{fig:inception} shows a screenshot of INCEpTION interface for \textsc{Assessment and Plan Relation Labeling} task. Linebreaks are naturally preserved from MIMIC-III raw data. On every line of text that belongs to Assessment and Plan sections (text spans marked with orange tags \textsc{A/P}), we further categorized them as \textsc{Assessment} or \textsc{Plan Subsection} (the blue color box above tag \textsc{A/P}). Then we linked the two text spans and marked the relation type (the tag \textsc{Direct} and blue dotted lines connecting the two text spans). A pop-up window would appear to show the linkage between the Plan Subsection and the Assessment.      

Two medical students were recruited as annotators. The medical students had completed their first year of medical school and were trained in the SOAP note format. They completed basic training in pathology, anatomy, pharmacology, and pathophysiology and also trained for an additional three weeks with the physician-scientists supervising the task. The inter-annotator agreement using Cohen's kappa score was measured between each annotator and the physician-scientist supervisor until each annotator met a threshold of 0.80 Kappa score with the supervisor. The supervisor audited the annotations periodically to re-assess the agreement would re-train if the kappa score fell below 0.80. During annotation, annotators augmented their knowledge with subscription and access to a large medical reference library, UpToDate®\footnote{\url{https://www.wolterskluwer.com/en/solutions/uptodate}}. The final inter-annotator agreement between the two medical student annotators was reported as 0.76, which was acceptable given the complexity of the task.

\begin{figure*}
\small
\centering
    \begin{tabular}{l|l} \hline 
    Criterion & Label \\ 
    \hline 
       Assessment section includes a primary diagnosis/problem and  &  \textsc{direct} \\
       it is mentioned in the Plan subsection.  \\ \midrule
       Progress note includes a primary diagnosis/problem for   & \textsc{direct} \\
       hospitalization and it is mentioned in the Plan subsection. &  \\ \midrule
       Plan subsection contains a problem/diagnosis related to  & \textsc{direct} \\
       the primary signs/symptoms in the Assessment section. &  \\ \midrule 
       Plan subsection contains complications/subsequent events   & \textsc{indirect} \\ 
       or organ failure related to the primary diagnosis/problem  & \\
       from the Assessment section. & \\ \midrule 
       Plan subsection contains other listed diagnoses/problems & \textsc{indirect}  \\ 
       from the overall Progress Note or in the Assessment & \\
       section that are not part of the primary diagnosis/problem. &  \\  \midrule 
       Plan subsection contains a diagnosis/problem that is not & \textsc{indirect} \\
       previously mentioned but closely related (i.e., same organ &  \\
        system) to the primary diagnoses/problems mentioned in   &  \\
       the overall Progress Note or Assessment section. &  \\  \midrule 
       None of the criteria for Directly Related or Indirectly & \textsc{neither} \\
       Related are met but a diagnosis/problem or other signs/ & \\ 
        symptoms are mentioned. &  \\ \midrule 
       Plan subsection does not include a diagnosis/problems OR & \textsc{not rel} \\
       signs/symptoms.  \\ 
       \hline
    \end{tabular}

    \caption{Guidelines for annotating the four relations (\textsc{direct, indirect, neither, not relevant}) between Assessment and each subsection of Plan}
    \label{fig:guidelines}
\end{figure*}

The final annotated corpus contained 768 progress notes and 5934 labels for the four relations. Specifically, we annotated 1404, 1599, 1913, and 1018 relations for \textsc{Direct, Indirect, Neither} and \textsc{Not Relevant}, respectively. 

\subsection{Task setup, timeline and evaluation }

\begin{figure}[]
    \centering
    \begin{tabular}{l|l} 
    \toprule 
         Assessment & Mr. [**Known lastname 512**] is a 60 y/o male with  \\
         & HTN who presents with a food impaction after ingestion  \\ 
         & of roast beef at the RedSox game.  \\ \midrule
         Plan Subsection & Opacity on CXR. Concerning for aspiration \\
         & pneumonitis. \\
         &   - Continue clindamycin for now \\ 
         &   - Will monitor patient for new O2 requirement, fever. \\  \midrule
         Relation & Indirect \\ 
         \bottomrule 
    \end{tabular}
    \caption{An example pair of Assessment and Plan Subsection with \textsc{Indirect} as the ground truth labels. ``Aspiration pneumonitis'' is the diagnosis mentioned in the Plan Subsection. }
    \label{fig:error_example}
\end{figure}

Figure~\ref{fig:error_example} presents an input example with the ground truth label as \textsc{Indirect}. To correctly predict the relation, a model needs to understand the medical concepts in both Assessment and Plan Subsections, and establish the connections between concepts given the input context. Each sample was organized as a pair of Assessment and Plan Subsection with the gold relation labels into csv files. As a result, we obtained 4670, 594, and 677 samples in training, development, and test set, respectively. Using a BERT tokenizer~\cite{devlin2019bert}, the average length of the Assessment section was 77.43 tokens, and the average length of each Plan Subsection was 86.84.  Along with the sample, we released the Hospital Admission ID (\textsc{HADMID}) indicating the particular admission the progress note was generated from. The N2C2 participants had the opportunity to use the entire progress note using \textsc{HADMID} as well as other note types, and we asked the participants to report if they used other note types as input to the model when they submitted their systems.  

The training set was released at the beginning of May, 2022 and the test set was released in mid-July 2022. Participants had two months to develop models. Each team was allowed to submit at most 9 runs, and we picked the best-performing system as the final performance for the teams. Given that the relation classes were distributed evenly, the Macro F1 score and the 95\% confidence intervals (CIs) were reported as the evaluation metric used for ranking systems.

During the final submissions of systems, each team was asked to complete a survey about system development. The survey addressed questions about the methods, external datasets, expert resources, input data, long document usage with the entire progress note, and multi-modality EHR usage.

\section{Results}
\subsection{Participation}
\begin{table}[ht]
\small 
    \centering
    \begin{tabular}{l|l|l|r} \toprule
       Team  & Institutions & Country & \# Runs  \\ \toprule 
      Carnegie Mellon University & Carnegie Mellon University  & USA & 9 \\ 
      (CMU)  & National Library of Medicine &  & \\ \midrule 
       \multirow{1}{*}{ University of Florida (UFL)} & University of Florida & USA & 4 \\ \midrule 
       \multirow{2}{*}{ Yale University (Yale)} & Yale University, &USA &  1\\ 
       & University College Dublin & Ireland \\ \midrule 
       University of Massachusetts& UMass-Amherst, & USA & 5\\ 
       - Amherst (UMass) & UMass-Lowell  & \\  \midrule 
       John Hopkins University (JHU) &  Harvard University, & USA & 3 \\  & John Hopkins University, &  \\ 
       & University of Pittsburgh & 
       \\ \midrule
       University of Wisconsin (UWisc) & University of Wisconsin & USA & 3\\ \midrule 
       University of Duisburg-Essen  & University of Duisburg-Essen & Germany & 3\\ 
       (UDuisburg) & & \\  \midrule 
       Dalian Minzu University  & Dalian Minzu University & China & 1  \\ 
       (Dalian-Minzu) & Dalian University of Technology & & \\ 
       \bottomrule 
    \end{tabular}
    \caption{Overview of participated teams: the institutions the teams represented, countries represented, and number of submitted runs. }
    \label{tab:participation}
\end{table}

Participating teams were required to sign a data use agreement to get access to the shared task. During the registration period, we had 53 participants registered for the shared task. Twenty-six participants across eight teams finished the model development and made successful submissions during the system evaluation period. Table~\ref{tab:participation} presents the teams, countries represented, and the number of submitted runs. 

\subsection{Team performance}

\begin{table}[ht]
\small 
    \centering
    \begin{tabular}{l|l|l} \toprule
       Rank  & Teams & Macro F1  \\ \toprule 
        1 & CMU & 0.8212  \\ 
2 & Yale &  0.8133   \\ 
3 &  UWisc & 0.8119  \\ 
4 & UDuisburg &  0.8034  \\
5 & JHU & 0.7994 \\ 
6 & UMass & 0.7949 \\ 
7 & UFL &  0.7947 \\ 
8 & Dalian-Minzu & 0.7454 \\
       \bottomrule 
    \end{tabular}
    \caption{Team performance ranking based on the best system performance.}
    \label{tab:best_ranking}
\end{table}

\begin{figure}
    \centering
    \includegraphics[scale=0.5]{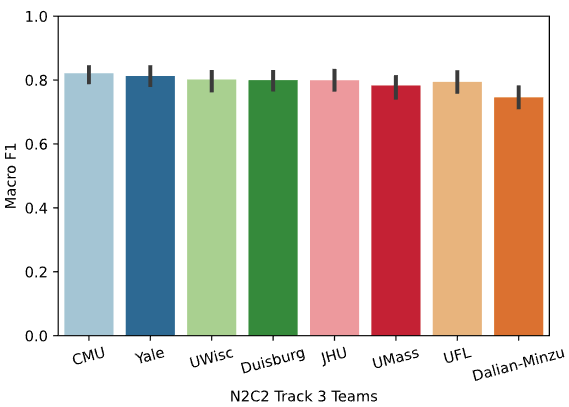}
    \caption{Macro F1 with 95\% confidence intervals from all teams. }
    \label{fig:CI_best}
\end{figure}

\begin{table}[ht]
\small 
    \centering
    \begin{tabular}{l|l|l|l|l|l} \toprule
       Teams  & Mean F1 & Median F1 & Min F1 & Max F1 & SD F1  \\ \toprule 
       CMU &  0.8064 &	0.8090	& 0.7920 &	0.8212 & 0.0099  \\ 
       Yale & 0.8133 & 0.8133 &	0.8133 & 0.8133	 &0.0000 \\
       UDuisburg & 0.8006 &	0.7998 &	0.7987	& 0.8034 & 0.0025 \\ 
       UWisc & 0.7990 &	0.7979 &	0.7872 &	0.8119 &	0.0124 \\
       UFL & 0.7915	& 0.7937 &	0.7839 &	0.7947 & 0.0051 \\ 
       JHU & 0.7851 &	0.7896 & 0.7663 & 0.7994 &	0.0170 \\ 
       UMass & 0.7833&	0.7785	& 0.7771 &	0.7949 & 0.0082 \\
       Dalian-Minzu & 0.7454	& 0.7454 &	0.7454 &	0.7454 &	0.0000 \\ \midrule
       All & 0.7949 &	0.7949 & 0.7454 & 0.8212 &	0.0160 \\ 
       \bottomrule 
    \end{tabular}
    \caption{Statistics over all submitted runs (scores reported in Macro F1).}
    \label{tab:average_ranking}
\end{table}

We received 29 submissions of system output and ran an evaluation script to generate the final scores. In this section, we presented the system ranking and average performance across all submitted systems. 

Table~\ref{tab:best_ranking} presents the ranking based on the best systems submitted by each team. The best optimal results were achieved by CMU with a Macro-F1 score of 0.8212, followed by the team Yale with an F1 score of 0.8133, and the system from UWisc followed with an F1 score of 0.8119.

\begin{table}[ht]
\small 
    \centering
    \begin{tabular}{l|l|l|l|l} \toprule
       Teams  & Direct & Indirect & Neither & Not Relevant  \\ \toprule 
       CMU &  0.8306 &	0.6917	& 0.8254 &	0.9372   \\ 
       Yale & 0.8328 & 0.6764 & 0.8158 & 0.9307 \\
       UWisc & 0.8101 &	0.7005 &	0.8227 &	0.9143 \\
       UDuisburg & 0.8038 &	0.6792 &	0.7999	& 0.9306  \\ 
       UFL & 0.8202	& 0.6593 &	0.8115 & 0.8878  \\ 
       JHU & 0.8115 &	0.6772 & 0.7972 & 0.9117 \\ 
       UMass & 0.8011 &	0.6667	& 0.8054 &	0.9064 \\
       Dalian-Minzu & 0.7391	& 0.6196 &	0.7229 &	0.9000 \\ \midrule
       All &  0.8062 & 0.6713 & 0.8001 & 0.9154 \\ 
       \bottomrule 
    \end{tabular}
    \caption{Macro F1 on the four label classes across the eight teams' best systems.}
    \label{tab:label_f1}
\end{table}

We ran an evaluation on best system prediction using bootstrapping on 10000 samples and plotted the 95\% CI in Figure~\ref{fig:CI_best}. The 95\% CI from the top 7 teams overlapped, indicating that there might not be a significant difference in the team performances. In Table~\ref{tab:average_ranking}, we presented the overall statistics on Macro F1 across all submitted systems. The team Yale achieved the highest average F1 score at 0.8133, followed by team CMU at 0.8064 and UDuisburg at 0.8006. The average performance across 29 submitted systems was 0.7949, with a standard deviation of 0.0160.

Table~\ref{tab:label_f1} presents the broken-down Macro F1 scores on the four labels across each team's best system. On the four labels, all teams achieved the highest performance on predicting \textsc{Not Relevant} with average scores as 0.9154, and most teams had \textsc{Direct} labels predicted correctly with average scores at 0.8062. \textsc{Indirect} is the hardest label to predict and reported as 0.6713 average macro F1. Finally, the average score on \textsc{Neither} is 0.8001.

\subsection{Methods overview}

\begin{table}[ht]
\small 
    \centering
    \begin{tabular}{l|l|r} \toprule
       Fields  & Explanations  & \# Yes\\ \toprule 
       method-used & Describe the methods. & - \\ 
        other-dataset & Did you use any other datasets besides MIMIC?  & 0  \\ 
        external-resource & Did you use any external medical resource? & 3 \\ 
        md-involved & Did you use medical doctors' expertise?  & 4\\ 
        additional-data & Did you use additional data as input?  & 4  \\ 
        entire-progress-note & Did you use the entire progress notes ? & 4\\ 
        other-parts & Did you use other parts of the notes? & 7 \\ 
        multi-modal & Did you use multi-modality?  & 0 \\  
       \bottomrule 
    \end{tabular}
    \caption{Fields in the submission form that describe the method development and the collected number of ``Yes'' over 29 submissions.}
    \label{tab:system_spec}
\end{table}

\begin{figure}
    \centering
    \includegraphics[width=\textwidth]{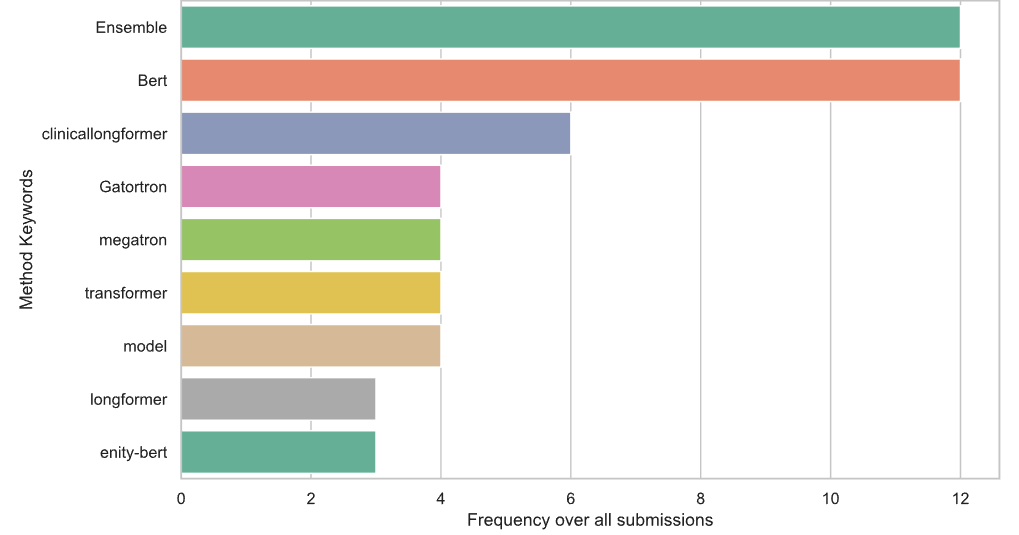}
    \caption{Top 10 most frequent keywords occurred in the submission descriptions. }
    \label{fig:most_freq_keywords}
\end{figure}

\begin{table}[]
    \centering
    \begin{tabular}{c|c}
    \toprule
      Transformer Type &	Count \\ \midrule
      BERT &	12 \\ 
    ClinicalLongformer &	6 \\ 
    GatorTron	& 4 \\ 
    Longformer	& 3\\ 
    EntityBERT	& 3 \\ 
    BioClinicalBERT & 3 \\ 
    T5	& 1 \\ 
    RoBERTa	& 1 \\ 
    BioClinicalLongformer	&1 \\  
    \bottomrule
    \end{tabular}
    \caption{Frequency of various Transformer models in submission. }
    \label{tab:transformers}
\end{table}


\begin{sidewaystable}[]
\adjustbox{max width=\textwidth}{%
\small 
    \centering
           \begin{tabular}{l|l|l|l|l|l|l} \toprule 
       Team  & Methods & external-resources & md-involved & additional-data & entire-progress-notes & other-parts  \\ \midrule 
         CMU & The team used ClinicalBERT ensemble without data shuffling.    & N & N & Y & N & Discharge Summaries \\
         & They applied Bayesian inference to combine posterior probabilities   &  & & & & \\ 
         & of single model output.  &  & & & & \\ 
         \midrule
         Yale & The team used RoBERTa-Large as base model and a Human-in-the-loop  & N & Y & N & N & N \\
         & approach  where clinicians annotated clinical notes for primary and   &  & & & &\\
         &  secondary problems. They then trained an NER tagger on the       &  & & & & \\ 
         & augmented data and applied the tagger with the base model for final  &  & & & &    \\ 
         & prediction. &  & & & &    \\  \midrule
         UWisc & They encoded concept features using International Classification of   & UMLS, MetaMap & N & N & N &N   \\ 
         & Diseases (ICD) codes provided by MetaMap. The final input included     & & & & & \\
         & concept features and text from Assessment and Plan Subsection and     & & & & & \\
         & was fed into an ensemble model with Enity-BERT and LightGBM. & & & & & \\ \midrule
         UDuisburg & They tested with general domain BERT model and ClinialBERT. & NER tagger, Stanza & N & N & N & N \\
         & For ClinicalBERT model, they utilized plan ordering feature and  & & & & &\\
         & Stanza NER tagger. The best system is an ensemble method over   & & & & &\\
         & CRF,  general domain BERT and ClinicalBERT.  &  &  &  &  \\ \midrule
         JHU & The team uses ClincalBERT and fine-tuned different clinicalBERT   &  N & N & N & N & N  \\ 
         & models by formatting the input Assessment and Plan Subsection as   & & & & &\\  
         & next sentence prediction task. Then they built ensemble model to  & & & & &\\
         &  make the final prediction using majority voting.  & & & & &\\ \midrule 
         UMass & The team utilized UMLS semantic type features, and built a ensemble & UMLS, MedCAT & Y & N & N & N \\ 
         & method over BioClinicalLongformer, ClinicalLongformer and LSTM. & & & & &\\
         & The system made prediction over sequence of input paragraphs. & & & & &\\ \midrule 
         UFL & The team utilized a pretrained transformer model – GatorTron, & N & N & N & N & N \\
         & which was developed using over 80 billion words of clinical narratives & & & & &\\ 
         & at UF Health using the BERT architecture. & & & & &\\ \midrule  
         Dalian-Minzu & The team reformulated the task as a masked language text & N & N & N & Y & N  \\ 
         & generation task by constructing cloze-style prompt
templates, & & & & &\\ 
& and fed the prompt and input text to general domain T5 models. & & & & &\\  
         \bottomrule
    \end{tabular} 
 }
    \caption{Best system and configurations from each team. }
    \footnote{The column headers of the table corresponding to the survey we collected from shared task submissions. In particular, \textit{external-resources} indicates if the team uses external resources (such as UMLS, MetaMap); \textit{md-involved} indicates if the team has a medical expert involving in system development and data analysis; \textit{additional-data} indicates if the team uses any data besides MIMIC;  \textit{entire-progress} indicates if the team the entire progress notes during pre-training or continuous training; \textit{other-parts} indicates if the team uses other parts of the notes or outside the notes for training. }
    \footnote{Abbreviations in methods: NER: Named Entity Recognition; UMLS: Unified Medical Library System.  }
    
    \label{tab:best_system_config}
\end{sidewaystable}

We received 29 submissions for the survey with the response rate for each question shown in Table~\ref{tab:system_spec}. None of the submissions used other datasets besides MIMIC or involved multi-modal models. Four teams answered ``Yes'' to the question of whether the entire progress note was being used or other parts of the notes were used; however, the additional sections or notes were only utilized in pre-training and continuous training. Four systems used discharge summaries as input, and four consulted medical doctors' expertise during system development. There were three systems that used external medical resources such as the National Library of Medicine's Unified Medical Language System (UMLS). 

We identified the keywords used in the ``method-used'' and plotted the top 10 most frequent words used in the submitted abstract methods (Figure~\ref{fig:most_freq_keywords}). We found ``Ensemble'' and ``BERT'' was mentioned 12 times across all submissions. All teams used transformer-based pre-trained language models~\cite{vaswani2017attention}, and concatenated each input Assessment and Plan Subsection with special tokens like ``[CLS]'' and ``[SEP]''. A summary of the full list of language models used is in Table~\ref{tab:transformers}. The original BERT model pre-trained on the general domain was used most frequently in ensemble methods with other BERT variants and machine learning algorithms (Bayes rules, Conditional Random Fields, etc.). BERT-based models that were pre-trained or continuously trained on the biomedical domain were also common in this shared task. For instance, GatorTron was trained on a corpus with over 90 billion words from the University of Florida Health System's EHR, PubMed articles, Wikipedia and MIMIC ~\cite{yang2022gatortron}. EntityBERT was continuously trained on PubMed and MIMIC dataset using medical entity masking strategy~\cite{lin2021entitybert}. Besides BERT, ten submitted system outputs were produced from Longformer~\cite{Beltagy2020Longformer} and its biomedical domain variants (``ClinicalLongformer''~\cite{li2022clinical}, ``BioClinicalLongformer''). Longformer is a BERT-variant with longer token limits and was designed to process document-level information, and it was a popular choice among all teams. 

Table~\ref{tab:best_system_config} summarized each team's best system design and configuration. Three of eight teams utilized the Unified Medical Language System (UMLS), a large medical vocabulary for medical concepts~\cite{bodenreider2004unified}.  The best-performing system developed by team CMU designed an ensembled BERT method and trained the models on both progress notes and discharge summaries. They utilized the Bayes Inference on different models. The second-best system developed by team Yale trained a NER tagger based on the training and development set, and a Human-in-the-loop approach with the help of two physicians to generate augmented data for training. Specifically, they have clinical experts labeled primary and secondary problems/symptoms, and complications from in-house ICU SOAP notes. The UWsic constructed a pipeline that first encoded hand-crafted features and disease and symptoms categories identified from a concept extractor (MetaMap~\cite{aronson2006metamap}), then fed these features into Entity-BERT models, running a LightBGM algorithm~\cite{ke2017lightgbm} to select the final prediction. Like team UWsic, team Duisburg employed both hand-crafted features and machine learning algorithms, specifically utilizing a Conditional Random Field (CRF) classifier and features from Stanza i2b2 clinical NER tagger~\cite{zhang2021biomedical} in BioClinicalBERT~\cite{alsentzer2019publicly} models. They took majority vote over ensemble BERT models as the final prediction, and yielded competitive performance. Team JHU also employed an ensemble method, specifically utilizing a majority vote over BioClinicalBERT models. Team UMass utilized a set of Longformer models including ClinicalLongformer~\cite{li2022clinical} and BioClinicalLongformer. Their best-performing system applied MedCAT~\cite{Kraljevic2021-ln} to identify the medical concepts as part of the input and a layer of Long short-term memory (LSTM)~\cite{hochreiter1997long} for final relation prediction. Team UFL ran GatorTron, a transformer-based model pre-trained on over 90 billion words including clinical and general domains. Finally, team Dalian-Minzu was the only team that used generative language models for this task. Specifically, they used T5~\cite{raffel2020exploring} with cloze-style prompt templates to generate the relations.    

We also found that two teams used Plan Subsection ordering information: team UWisc and team UDuisburg, but the effects of incorporating such information was not significant.  

\section{Discussion}
 
For Track 3 of the n2c2 shared task, we attracted international participation and eight teams with 27 participants contributed to solving the task. The best system achieved 0.8212 Macro-F1 score, and the average Macro F1 score across all submitted teams was 0.7949. All the submitted systems were developed based on BERT and Transformer methods, demonstrating the impact that pre-trained language models have had on the field of clinical NLP. We also observed that the most common errors occurred on \textsc{Indirect} and \textsc{Neither} predictions possibly because the useful information to determine these two relations were contained in other sections of the progress note and required complex medical reasoning. None of the tasks used the full progress note and the different modalities of data from the SOAP sections (overnight events, vital signs, physical exam findings, laboratory results, iamging results), so future research may include working on long document processing and  multi-modal cNLP models. 

Most ensemble methods performed better than single models, and combining hand-crafted features from knowledge sources with observations of the data helped the models predict relationships with greater accuracy than without. The major medical knowledge source for pre-trained models included EHR data, PubMed articles, and Wikipedia. We observed that models trained on EHR data showed a subtle advantage over the other pre-trained models, as two of the top 3 systems were trained on MIMIC dataset. In particular, the top-performing system showed that utilizing other EHR note types (i.e., discharge summaries) to augment training data and finding posterior estimates with Bayes Inference over ensemble BERT models helped to augment the models' performance. We also observed that applying NER tagging on the input text could help the model better capture the relations between concepts. In particular, team Yale demonstrated that a model benefited from the human-in-the-loop approach where clinical experts would identify the primary and secondary diagnoses from the input data and inject the domain knowledge into pre-trained language models. Their approach achieved competitive results by being the second-top-performing system.  

Several teams observed from the training data that the \textsc{Direct} relations often occurred at the beginning of the Plan sections. They developed models that incorporated the ordering information of Plan Subsections but the improvements were not significant. The primary diagnosis may be placed before other diagnoses in some physicians' writing, but this was not always the case and the ordering of that Plan section proved not to be helpful.     

Longformer was a popular model selection among all teams, but the improvements over BERT models with smaller token limits did not provide much performance gain, possibly due to few samples exceeding the token limit. If participants had used the entire progress note then Longformer would have been more applicable to ingest the long document of daily care notes. During annotation, full access to the progress note was provided and the full document was used in reasoning for the labels. Modeling the behavior of the annotators in the NLP models was not an approach taken by any participants. Important details were available in the Subjective sections and Objective sections that were missed in all model inputs. For instance, the diagnosis of ``leukocytosis'' may be reflected in blood test results of the Objective section of the SOAP progress note. Understanding the blood test results would also require the system's capacity in understanding the structured text. Further, some diagnoses required complex medical reasoning. Figure~\ref{fig:error_example} presents an example Assessment and Plan Subsection with the ground truth relation as \textit{Indirect} where all systems predicted incorrect labels. To diagnose a patient with aspirations pneumonitis, typically a healthcare provider would perform imaging and other diagnostic tests. Although aspirations pneumonitis is not directly related to hypertension (``HTN'' in the Assessment section in Figure~\ref{fig:error_example}), hypertension was a mentioned comorbidity that may confer a higher risk for a poor health outcome from the primary diagnosis of pneumonitis. Further, our task surfaces the potential role for a multi-modal NLP model with advanced medical reasoning that could improve the performance by understanding and interpreting the clinical evidence from other parts of the progress note that include vital signs, physical exam findings, laboratory data, and imaging results to infer logical relations on the concepts that were distantly related.   

Several limitations occurred in the shared task. First, no formal sample size calculations were performed to identify the number of labels and notes needed for a well-powered task.  The small differences in F1 scores between participants may have been attributable to similar use of pre-trained language models but also that the sample size in the test set was underpowered to detect a difference. Future directions in shared tasks should include sample size calculations a priori.  Further, our sampling was performed from a single health system and was limited to ICU progress notes. While the SOAP format is ubiquitous, variations in notetaking style by discipline and templated language from diferent EHR vendors likely provide more heterogeneity in progress notes than was available in  MIMIC. Multi-center sampling across more disciplines is another future direction to further challenge model performance.

\section{Conclusion} 
Overall, the top-performing models achieved high F1 scores but room for improvement remains. This task addressed a specific clinical use that helps providers to prioritize their time on the most urgent diagnoses and health problems. Yet, our single task is not enough to transform the focus of the cNLP field to promote model developments for clinical applications that assist clinical decision-making. For an NLP-driven clinical decision support system that could be run at the bedside, the basic design decisions will require NLP models that are capable of utilizing medical knowledge and understanding and synthesizing information. In the suite of cNLP tasks where this shared task originated from, two additional tasks are publicly available to facilitate model development and evaluation for clinical diagnostic reasoning~\cite{gao2022tasks, goldberger2000physiobank}. The two other tasks are a SOAP Section Labeling task and a Problem Summarization task, and both tasks are widely recognized as useful tasks for note-writing practice and diagnostic decision-making~\cite{gao-etal-2022-hierarchical}. We thus encourage the community to create, follow and conduct research on clinical diagnostic reasoning and advance the field of cNLP.

\appendix
\section*{Data Availability}
 ``MIMIC-III' is available at PhysioNet (\url{https://physionet.org/content/mimiciii-demo/1.4/}).
 The data used in this N2C2 challenges is available at N2C2 website. (\url{https://n2c2.dbmi.hms.harvard.edu/}).

\section*{Competing Interest}
No competing interest is declared. 

\section*{Declarations}
The research data used in this work is only available through PhysioNet and the original publication. Data Use Agreement (DUA) is required for MIMIC-III based dataset. We do not claim authorship over the dataset.  

\section*{Funding}
The work was supported by NIH/NIDA grant number R01DA051464 (to MA), NIH/NIGM grant number R01HL157262 (to MMC), NIH/NLM grant numbers R01LM012793 (to TIM), NIH/NLM grant number R01LM010090 (to DD), NIH/NLM grant number R13LM013127 (to OU). The content is solely the responsibility of the authors and does not necessarily represent the official views of the National Library Of Medicine or the National Institutes of Health.

 \bibliographystyle{elsarticle-num} 
 \bibliography{bibs}
 

\end{document}